\documentclass[12pt]{article}
\usepackage{algorithm}
\usepackage{algpseudocode}
\usepackage{amsmath}
\usepackage{amsfonts}
\usepackage{graphicx}

\usepackage{natbib}
\usepackage{setspace}
\usepackage{url} 
\usepackage{framed}
\usepackage{comment}
\usepackage{algorithm}
\usepackage{algpseudocode}
\usepackage{hyperref}
\usepackage{subcaption}
\usepackage{multirow}
\usepackage{comment}
\usepackage{amsmath}
\usepackage{makeidx}
\usepackage{xr}
\newcommand{\blind}{0}
\begin{document}

\def\spacingset#1{\renewcommand{\baselinestretch}%
{#1}\small\normalsize} \spacingset{1}


\if0\blind
{
  \title{\bf Likelihood-Free Estimation for Spatiotemporal Hawkes processes with missing data and application to predictive policing}
  \author{Pramit Das, Moulinath Banerjee, Yuekai Sun  \hspace{.2cm}\\
    Department of Statistics, University of Michigan, Ann Arbor\\
  }
  \maketitle
} \fi

\if1\blind
{
  \bigskip
  \begin{center}
    {\LARGE\bf Likelihood Free Estimation for Hawkes processes with missing data with application to predictive policing}
\end{center}
  \medskip
} \fi

\bigskip

\begin{doublespace}
\begin{abstract}
\noindent With the growing use of AI technology, many police departments use forecasting software to predict probable crime hotspots and allocate patrolling resources effectively for crime prevention. The clustered nature of crime data makes self-exciting Hawkes processes a popular modeling choice. However, one significant challenge in fitting such models is the inherent missingness in crime data due to non-reporting, which can bias the estimated parameters of the predictive model, leading to inaccurate downstream hotspot forecasts, often resulting in over- or under-policing in various communities, especially the vulnerable ones. Our work introduces a Wasserstein Generative Adversarial Networks (WGAN) driven likelihood-free approach to account for unreported crimes in Spatiotemporal Hawkes models. We demonstrate through empirical analysis how this methodology improves the accuracy of parametric estimation in the presence of data missingness, 
leading to more reliable and efficient policing strategies.

\vspace{0.2cm}

\end{abstract}
\end{doublespace}

\begin{doublespace}

\section{Introduction to Hawkes Processes and Predictive Policing}
\label{sec:Intro to predictive policing in Bogota}
In the statistics and probability literature, spatiotemporal processes are critical tools in understanding how events are linked across time and space. While Non-homogenous Poisson Processes (NHPPs) are widely used for modeling spatial and temporal events, they are not well suited for situations with phenomena that exhibit a pattern of self-triggering or clustering.  
In such cases, the intensity is typically not only a function of time and space but also the spatiotemporal coordinates of past events. Such self-exciting Poisson processes (SEPP's) are called Hawkes processes and were introduced in the seminal paper \cite{hawkes1974cluster}. The intensity function $\lambda(\cdot)$.
is given by
\begin{equation}
\label{HawkesIntensity} \textstyle
\lambda(t,x,y|\mathcal{H}_t) = \mu_0(x,y) + \sum_{t_i < t}\,g(t-t_i,x-x_i,y-y_i ) \,,
\end{equation}

where $\mathcal{H}_t =\{t_i \mid t_i <t,\:i\in \mathbf{N}\}$ denotes the history of the process till time $t$, $\mu_0(x,y)$ represents the background (base) intensity and the kernel $g(t-t_i,x-x_i,y-y_i )$ describes the triggering effect of past events. These models have applications in various fields such as seismology (\cite{JASAearthquake}), finance (\cite{hawkes2018hawkesfinance}, \cite{bacry2015hawkes}) and neuroscience (\cite{neurosciencerossant2011sensitivity}), where the occurrence of an event increases the likelihood of future events in the neighborhood. Such patterns are often observed in crime data, so spatiotemporal Hawkes models have therefore been used to predict future crime hotspots via software such as PredPol ( \cite{UCLAcrimepaper}). In fact, with the rise of machine learning and AI tools, police departments increasingly deploy such software to analyze past crime data to identify future `hotspots' (high crime zones) to which increased patrolling resources will be allocated. This is often referred to as Predictive Policing. However, using these models raises concerns about civil rights, privacy, and potential biases that could reinforce over-policing in certain communities. Incorporating fairness, accountability, and transparency in the design and implementation of predictive policing models is essential to mitigate these risks and to ensure that they complement traditional policing methods effectively (\cite{NiljanaAkpinarpredpolicing_2021}, \cite{favour2024predictivepolicing}).
\newline

Our work in this paper is motivated by earlier analyses of a crime simulation model based on real (crime) data from Bogota, Colombia in \cite{NiljanaAkpinarpredpolicing_2021} and our proposed methods are demonstrated on a similar crime simulation model, so we describe it next.

\begin{figure}[h]

    \centering
   \includegraphics[width=10cm]{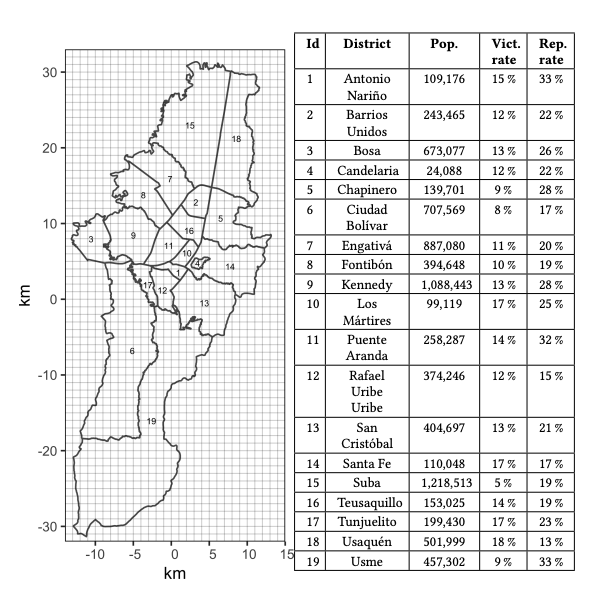}
    \caption{Bogota map and crime statistics: figure from \cite{NiljanaAkpinarpredpolicing_2021}}
  \label{fig:Bogota map}
\end{figure}

Bogotá, the capital of Colombia, has a population of approximately 7 to 8 million individuals across its 19 districts (see Figure \ref{fig:Bogota map}). We obtain district-specific crime reporting rates through a biannual survey conducted by the Bogotá Chamber of Commerce. (see \cite{NiljanaAkpinarpredpolicing_2021} for details). \cite{predictivepolicing} use certain summary statistics the Bogota crime data to understand the differential rate of crime reporting in various districts on downstream predictive policing tasks, such as predicting crime hotspots. Since real crime data is usually proprietary and accessible only by police departments, they resort to a simulation study using the Hawkes process framework, as considered by several previous authors, e.g, \cite{UCLAcrimepaper}, \cite{networkreconstructionSThawkes},\cite{YaoXiecrime911calltextAtlanta2020spatial}.  
To simulate crimes occurring in Bogota, \cite{predictivepolicing} use a spatiotemporal Hawkes model with background intensity:  

\begin{equation}
    \mu(t,x,y) =\mu \sum_{c_i}  \frac{1}{2\pi \sigma_0^2}\exp(-\frac{||(x,y)-c_i||^2}{2\sigma_0^2})
\end{equation}

where $c_i$'s are certain centers in the grid map, namely 
$$
\{(\pm6, \pm 20), (\pm 6, \pm 10), (\pm 6, \pm 30),(\pm 6,0)\}
$$ and $\sigma_0^2=4.5^2$ is a known number chosen according to the shape and size of the map and, therefore, not considered as a parameter in the subsequent analysis. The triggering kernel is a  product kernel with exponential decay in the temporal dimension and a Gaussian kernel in the spatial component given by:
$$
g(t, x, y)=\theta \omega \exp (-\omega t) \exp \left(-\frac{x^2}{2 \sigma_x^2}\right) \exp \left(-\frac{y^2}{2 \sigma_y^2}\right)
\,.$$

Note that, alternatively, it is possible to reparametrize the triggering kernel using $(\alpha, \beta) = (\theta \omega, \omega)$ and write the temporal part as $\alpha \exp(-\beta t )$.

\begin{figure}[h]
    \centering
   \includegraphics[height=7cm]{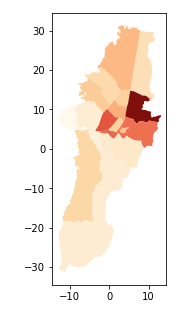}
    \caption{Bogota district-wise comparison of expected crimes}
    \label{fig:Bogota2}
\end{figure}
    
The crime data generation mechanism is as follows:
\begin{itemize}
    \item Generate events $C=\left\{\left(t_i,x_i, y_i,\right)\right\}_{i=1}^N$ from the Hawkes model above and discard all events outside the city limits. We denote by $N$ the number of events that occur over the time horizon of the study, denoted by $T$. 

    \item For each of the $19$ districts $d$, denote the data within its bounds by $C_d \subseteq C$. Subsample $n_d \sim \operatorname{Bin}\left(\left|C_d\right|, p_d\right)$ of the points to form the true crime data set $\mathcal{D}$, where
$$
p_d=\frac{\operatorname{population}(d) \cdot \text { victimization } \operatorname{rate}(d) \cdot \frac{T}{\text{6 months}}}{\left|C_d\right|}\,.
$$
Note that this resampling only has an effect if $p_d <1$ for a given district $d$. The idea behind it is to ensure that in the long run the crime numbers in each district align with the numbers expected from the population and the half-yearly victimization rates (available from the table). The retained set of crimes in district $d$ is denoted by $ \mathcal{D}_d$, where $\left|\mathcal{D}_d\right| =n_d$.

\item  For further thinning to a data set of only reported crimes, subsample $m_d \sim$ $\operatorname{Bin}\left(\left|\mathcal{D}_d\right|, q_d\right)$ crimes for each district $d$ where
$q_d=\text {crime reporting } \operatorname{rate}(d)$.

\end{itemize} 

Note that this data generation process is an example of a spatiotemporal Hawkes model followed by thinning, and the likelihood of such a dataset cannot be tractably written. One central aspect of  \cite{NiljanaAkpinarpredpolicing_2021, predictivepolicing}'s study is to demonstrate the problematic consequences of downstream prediction tasks using parameter estimates obtained from running a maximum likelihood estimation on only the reported data \textit{assuming it to be a full Hawkes model}. 

They use $\mu=100$, $\theta = 15$, $\omega = 0.2$, $\sigma_x= \sigma_y = 0.1$ to simulate the crime data and optimize  
The resulting likelihood function using the expectation-maximization (EM) algorithm, which is customary in these problems (See \ref{sec:Hawkes EM algorithm} and \cite{reinhart2018review}). We note that the `likelihood function' in the previous sentence refers to the likelihood function from \emph{the full Hawkes model}, but even this cannot be readily optimized by direct methods.  
\vspace{0.25cm}
\\

Following \cite{NiljanaAkpinarpredpolicing_2021}, we estimate the parameters from \emph{reported crime data without accounting for missingness or gaps in the data}. The results below show the averages over 10 simulation runs spanning 6-year data cycles. 

\begin{table}[ht]
\centering

\begin{tabular}{ |p{2cm}||p{2.3cm}|p{2.3cm}|p{2.3cm}| p{2.3cm}|p{2.3cm}| }
 \hline
 \multicolumn{6}{|c|}{Estimates from Reported Crime Data} \\
 \hline
 $T$(in years) & $\hat{\mu}$ & $\hat{\theta}$ & $\hat{\omega}$ & $\hat{\sigma_x}$ & $\hat{\sigma_y}$ \\
 \hline
 $T=2$   & 4.905 & 0.894 & 0.039 & 0.2269 & 0.2148 \\
  \hline
 $T=4$   & 4.565 & 0.817 & 0.040 & 0.2224 & 0.2248 \\
  \hline
 $T=6$   & 4.453 & 0.832 & 0.0394 & 0.228 & 0.229 \\
 \hline
\end{tabular}
\caption{Parameter estimates obtained using the EM algorithm applied to reported crime data. The table reports the mean estimates from 10 simulation runs.}

\label{tab:Bogota_EM_estimates_table}
\end{table}

The estimates of $\mu$, $\theta$, and $\omega$ obtained from the reported crime data are significantly different from the true parameter values used to simulate the data: ignoring the missing data produces gross underestimates of $(\mu, \theta, \omega)$. Consequently, downstream tasks, such as predicting crime hotspots, based on these estimates will produce highly biased predictions of the number of future crimes.  \\

It is clear that if we are to use maximum likelihood techniques to get consistent parameter estimates, we have to optimize the likelihood function for the \emph{missing data}, not the full Hawkes likelihood, as was done above. However, the \emph{missing data likelihood} is \emph{even more intractable} since it involves integrating out the Hawkes likelihood over the distribution of \emph{all missing events}, and neither their locations nor their numbers are known.  To address this challenge, we propose bypassing the intractable log-likelihood problem by adopting a \textit{likelihood-free learning method} such as a Wasserstein GAN (\cite{arjovsky2017wasserstein}).
The WGAN framework allows us to sidestep the complexities of calculating the likelihood of missing data by learning the distribution of the observed data through a generative method. This approach ensures robustness to missingness while enabling parameter estimation that better aligns with the true underlying generative process. In the remainder of this paper, we develop a WGAN-based parametric estimation method to account for missingness in Hawkes models.

\end{doublespace}

\begin{doublespace}

\section{Likelihood-Free Learning for Stochastic Processes}

In this section, we introduce a method for likelihood-free learning, using tools from optimal transport and generative adversarial networks (GANs) to recover parameters from crime data with missing values. We begin with some background on generative adversarial modeling (\cite{GANoriginalpaper}), Wasserstein GAN (\cite{arjovsky2017wasserstein}), and sequential models like recurrent neural networks (RNNs) and long short-term memory (LSTM) networks (\cite{hochreiter1997originalLSTM}). We then demonstrate how these methods can estimate the Hawkes model parameters in the context of the Bogota crime data simulation.
\newline
\newline
Consider a temporal or spatiotemporal point process governed by a parameterized distribution \( \mathbf{P}_{\theta} \). The goal is to learn the mechanism \( \mathbf{P}_{\theta} \) or parameter \( \theta \) from sample paths. As noted above, traditional methods like maximum likelihood estimation cannot be used when the likelihood is intractable, e.g., without an amenable closed-form representation, as in many processes involving random noise or missingness. Rather than relying on likelihood, we can also assess model fit by comparing the similarity between the true and proposed distributions. This approach is aligned with the framework of minimum distance estimators (MDEs), where the parameters are estimated by minimizing the divergence between the empirical distribution of the observed data and the model-generated distribution (
\cite{MDE_note}). MDEs provide a flexible alternative to the traditional maximum likelihood-based estimation framework. A modern incarnation of MDEs is given by the GAN formulation which we now discuss. 

\subsection{Generative Adversarial Networks (GAN) and Wasserstein GAN (WGAN)}

Generative Adversarial Networks (GANs) (\cite{GANoriginalpaper}) can be interpreted as a modern, iterative implementation of the minimum distance estimation idea, and were introduced by \cite{GANoriginalpaper} to generate samples from complex probability distributions using an adversarial approach. GANs consist of a generator \( G \) that creates `fake' samples and a discriminator \( D \) that determines whether samples are real or fake. Both models are typically chosen to be neural networks, allowing the handling of complex data structures. GANs, leveraging modern hardware like GPUs and TPUs, excel in tasks such as image synthesis, video generation, and text-to-image conversion and are particularly suited for likelihood-free learning in cases where the likelihood is intractable.
\vspace{0.1cm}

The optimization problem in GANs is framed as a minimax game between the generator and discriminator, where \( G \) tries to generate samples indistinguishable from the real data, while \( D: \mathcal{X} \rightarrow [0,1] \), seeks to classify real versus fake data correctly. Mathematically, the objective is to minimize the  following expression:

\[
\min _\theta \max _D V(G_{\theta}, D) = \min _\theta \max _D E_{x \sim p_{\text{data}}(\cdot)}[\log D(x)] + E_{z \sim r(\cdot)}[\log (1 - D(G_\theta(z)))]
\]

where $V(G_{\theta}, D)$ is the so-called Jensen-Shannon divergence. Here, \( z \sim r(\cdot) \) represents a source of randomness, \( G_\theta(z) \) denotes a generated sample from the generator, and \( D(x) \) represents the probability that a sample \( x \) comes from the true data. Usually, $r(\cdot)$, which is a base source of randomness, is taken as a uniform or Gaussian distribution. 
\\

 However, training GANs can be challenging due to mode collapse and unstable gradients (\cite{ChallengesoftrainingGANDivyaSaxena}). To improve on standard GANs, Wasserstein GANs (WGANs) were introduced by \cite{arjovsky2017wasserstein} to address certain limitations, such as when the true data distribution lies on a lower-dimensional manifold, where Jenson-Shannon (JS) and KL divergence are problematic. 
 Instead, WGANs use the Wasserstein distance, which offers better stability in training and helps the generator converge more effectively. The WGAN loss is also more interpretable as it directly reflects the Wasserstein distance between the real and generated data distributions. Using Kantarovich duality, the WGAN framework boils down the original minimax problem to:

\[
\min_{\theta} \max_{\| f_w \| \leq 1} E_{x \sim p_{data}(\cdot)}[f_w(x)] - E_{z \sim p_z}[f_w(G_\theta(z))]
\]

where \( f_w(\cdot) \) is a discriminator constrained by the $1$-Lipschitz condition, namely, \(\| f_w \|_L \leq 1\), and \( G_\theta(z) \) is the generator, which typically remains model-agnostic.

\noindent Using the Wasserstein distance as the loss function in WGAN leads to better stability than Jenson-Shannon divergence (JSD) based GANs. It also gives a meaningful loss function that directly corresponds to the similarity of the generated data to the true samples and addresses common pitfalls of traditional GAN, such as mode collapse; see \cite{arjovsky2017wasserstein} for details. Due to these salient advantages, \cite{WGAN+TPP} used Wasserstein GAN's for temporal point process modeling through recurrent neural networks that handle sequential data inputs. Here, in our work, we also use the Wasserstein distance as the choice for the distance metric between \( \mathbf{P}_{r} \) and \( \mathbf{P}_{\theta} \), but as will be explained below our $G_{\theta}$ generators are no longer model agnostic and will be elaborated on later.

\subsection{Recurrent Neural Network and Long Short-Term Memory Networks}

 While feed-forward neural networks are useful for various tasks, they lack memory of previous data occurrences, making them unsuitable for sequential data analysis, particularly time series data like ours. Recurrent Neural Networks (RNNs) address this limitation by maintaining an internal hidden state, allowing them to learn relationships and patterns across time stamps. RNNs are particularly effective for processing sequential inputs, such as text data or spatiotemporal event data, making them suitable for sequence-to-sequence tasks or assigning scalar values to sequences. Though RNNs have been widely used for sequence-to-sequence modeling, they suffer from challenges like vanishing/exploding gradients and difficulty retaining information over long sequences. To address these issues, \cite{hochreiter1997originalLSTM} introduced Long Short-Term Memory (LSTM) networks, which use memory cells with three gates, namely Input, Forget, and Output gates, that regulate information flow. These models are more flexible in dealing with sequential data and have gained popularity in the deep learning literature; see, e.g., \cite{paddinginLSTM} for technical details and also  \ref{sec:LSTM+RNN_math} in the supplement for a brief exposition.

\subsection{Generative models for Temporal point processes}
\label{sec:Xiao's work}
With the rise of deep learning, there has been an interest in fitting flexible sequences to sequence models ( such as recurrent neural networks, LSTMs, etc.). We revisit  \cite{WGAN+TPP}'s work on training generative models for temporal point processes. 
\newline

Given a number of sample paths from a stochastic process, the key idea is to look for a generative model $\mathbf{P}_\theta$ which is close enough to the true distribution $\mathbf{P}_r$.  In traditional GANs, Gaussian or uniform distributions are used as a source of random noise to generate $z \sim r(\cdot)$, which goes into the generator as an argument to generate a sample $g_\theta(z)$. For point processes, the authors in \cite{WGAN+TPP} use a homogeneous Poisson process as the building block of randomness in the space of point processes. 

 This is, in a sense, analogous to a uniform distribution on $\mathbf{R}^d$, since the number of events in any given region of $\mathbf{R}^d$ for a homogeneous Poisson process is proportional to its Lebesgue measure. 
\newline 
\\
Let $\zeta \sim r(\cdot)$ denote a sample path from a homogenous Poisson process on the time horizon $[0, T)$ with the constant rate $\lambda_z>0$ and $g_\theta(\cdot)$ be an RNN/LSTM type sequence to sequence translator that gives the sequence $g_\theta(\zeta)$. In \cite{WGAN+TPP}, the authors consider $\lambda_z$ as a part of existing domain knowledge about the problem. Hence, using Kantarovich duality, the minimax problem can be represented as:
$$
\min _\theta \max _{w \in \mathcal{W},\left\|f_w\right\|_L \leq 1} \mathbb{E}_{\xi \sim \mathbb{P}_\text{data}}\left[f_w(\xi)\right]-\mathbb{E}_{\zeta \sim r(\cdot)}\left[f_w\left(g_\theta(\zeta)\right)\right]
$$

where $f_w(\cdot)$ is  the discriminator and $g_\theta(\cdot)$ is  the generator model. This method in  \cite{WGAN+TPP}'s work enables us to model datasets using a black box RNN type generator flexibly but doesn't provide any interpretable model parameters or inference. While black box models are useful, due to the recent emphasis on transparency and interpretability of ML methods, in many instances, practitioners want to fit a \textit{specific interpretable point processes model}, such as the Hawkes process, birth-death chains, or Poisson process, etc. rather than training a black box generator. To resolve this issue, next, we propose our methodology in \ref{sec:Our method}. 
\end{doublespace}

\begin{doublespace}
    \subsection{Proposed Wasserstein GAN-based method with exact interpretable model generators}
\label{sec:Our method}

As explained in the \ref{sec:Xiao's work}, our proposed approach is motivated in part by the work in \cite{WGAN+TPP}, where the authors propose an approach for modeling temporal point processes using WGANs. Our main contributions are:

\begin{itemize}
    \item In our work, we develop an extension of the WGAN-based modeling technique in \cite{WGAN+TPP} with \emph{ model-specific exact generators}. Unlike black box models, this can give us model-specific interpretable parametric estimates.

    \item We extend \cite{WGAN+TPP}'s work on temporal point processes to \emph{spatiotemporal modeling and also to missing data}. 

\end{itemize}
\noindent Suppose the training data consists of a number of IID sample paths from a given point process where the stochastic generation mechanism is characterized by a parameter $\theta$ (e.g., $\theta = (\mu, \alpha, \beta, \sigma^2)$ for a spatiotemporal Hawkes process) and the generated process can be written as $g_{\theta}(z)$ where $z$ is some base source of randomness from the distribution $r(\cdot)$. 
The key point is that, now,  $g_\theta(\cdot)$ \textit{is an exact generator that can be written in an explicit and amenable algebraic expression}. Examples include inhomogeneous Poisson processes, Hawkes processes, missing data Hawkes processes, mixture models, and others. 
\\

 \noindent 
 We use the Wasserstein GAN procedure for estimating/recovering the exact generator $g_{\theta}(\cdot)$ by estimating $\theta$. This is a fairly general setup and extends the work of \cite{WGAN+TPP} to a parametric recovery setup for spatiotemporal processes with missingness or distortion. As before, the corresponding optimization problem at the population level is: 
$$
\min_{\theta \in \Theta}\max_{||f_w|| \leq 1}E_{x \sim P_\text{data}}[f_w(x)]-E_{z \sim r(\cdot)} [f_w(g_{\theta}(z))]
$$

However, in practice, we solve the empirical version: 
$$
\min_{\theta \in \Theta}\max_{||f_w|| \leq 1} \frac{1}{L}\sum_{l=1}^{L} f_w(x_l)
-\frac{1}{L}\sum_{l=1}^{L} f_w(g_{\theta}(z_l)) 
$$\noindent where $z_1,z_2,..,z_L$ is an IID sample from $r(\cdot)$ and $x_1,x_2,...,x_L$ is a random sample from the training data.
\newline
 In our method, to enforce the 1-Lipschitizness of the discriminator $f_w(\cdot)$, we use the gradient penalty introduced by \cite{Gradientpenalty} for training Wasserstein GAN's and minimize the loss function:

$$
l= E_{z \sim r(\cdot)} [f_w(g_{\theta}(z))]  -E_{x \sim P_\text{data}}[f_w(x)] + \lambda E_{\hat x} [(||\Delta_{\hat x} f_w(\hat x)||-1)^2]
$$

\noindent where $\hat x$ is a linear combination of samples from real and fake batches, namely $\textbf{x}$ and $\tilde{\textbf{x}}$ as described in Section 4 of \cite{Gradientpenalty}. Once again, in practice, during the optimization loop, we use the mini-batch version of the penalized loss function for Wasserstein GAN with the gradient penalty. The empirical batch version of the loss is given by:
$$
\hat{l} = \frac{1}{L} \sum_{i=1}^L f_w(g_{\theta}(z_i)) - \frac{1}{L} \sum_{i=1}^L f_w(x_i) + \lambda \frac{1}{L} \sum_{i=1}^L \left( \|\nabla_{\hat{x}_i} f_w(\hat{x}_i)\|_2 - 1 \right)^2,
$$

Here, \( f_w(\cdot) \) represents the discriminator parameterized by \( w \), \( x_i \) denotes the real samples drawn from the distribution \(\mathbb{P}_r\), and $g_{\theta}(z_i)$'s are the generated samples drawn, where \( z_i \sim r(\cdot)\) is a source of randomness (for example, uniform or Gaussian). The interpolated samples \(\hat{x}_i\) are defined as 
\[
\hat{x}_i = \epsilon x_i + (1 - \epsilon) \tilde{x}_i,
\]
where \(\epsilon \sim U[0, 1]\) is sampled uniformly from the interval \([0, 1]\).  Here, the hyperparameter \(\lambda\) controls the strength of the gradient penalty, while \( L \) denotes the batch size. We outline the WGAN training process with gradient penalty for spatiotemporal processes in the Algorithm \ref{alg:our_algo_flowchart} below.  

\begin{algorithm}[!ht]
\caption{WGAN with gradient penalty. We use default values of \( \lambda = 10 \), \( n_{\text{critic}} = 5 \), ADAM hyperparameters \( \alpha = 0.0001 \), \( \beta_1 = 0 \), \( \beta_2 = 0.9 \).}
\label{alg:our_algo_flowchart}
\begin{algorithmic}[1]
\Require Gradient penalty coefficient \( \lambda \), number of critic iterations per generator iteration \( n_{\text{critic}} \), batch size \( L \), initial critic parameters \( w_0 \), initial generator parameters \( \theta_0 \), Adam hyperparameters \( \alpha, \beta_1, \beta_2 \).
\While{the empirical loss has not converged}
    \State \textbf{Step 1: Generate fake samples}
    \State Generate a sequence of random vectors \( z=\{z_i\}_{i\geq 1} \), where \( z_i \sim r(\cdot) \).
    \State Generate \( L \) spatiotemporal sequences \( \{g_{\theta}(z_j) = \{(t_i^j,x_i^j,y_i^j)\}_{i \geq 1}\} \) for \( j = 1, \dots, L \).
    \State Represent the set of generated samples as \( S_{\theta} \), a tensor of suitable dimensions.

    \State \textbf{Step 2: Updating the discriminator \( D \)} 
    \For{\( t = 1 \) \textbf{to} \( n_{\text{critic}} \)}
        \For{\( j = 1 \) \textbf{to} \( L \)}
            \State Sample real data \( \zeta_j \sim P_r \) and a uniform random variable \( \epsilon \sim U[0,1] \).
            \State \( \tilde{S}_{j} \gets \epsilon \zeta_j + (1 - \epsilon) S^{(j)}_{\theta} \)
            \State Compute loss:
            \[
            l^{(j)} \leftarrow f_{w_{t-1}}(\tilde{S}_j) - f_{w_{t-1}}(S^{(j)}_{\theta}) + \lambda \left(\left\|\nabla_{\tilde{S}_j} f_{w_{t-1}}(\tilde{S}_j)\right\|_2 - 1\right)^2
            \]
        \EndFor 
        \State Update \( w \) using Adam:
        \[
        w \gets \text{Adam}(\nabla_w \frac{1}{L} \sum_{j=1}^L l^{(j)}, w, \alpha, \beta_1, \beta_2)
        \]
    \EndFor

    \State \textbf{Step 3: Updating the generator \( G \)}
    \State Sample a batch of latent variables \( \{z^{(j)}\}_{j=1}^L \sim r(\cdot) \).
    \State Update \( \theta \) using Adam:
    \[
    \theta \gets \text{Adam}(-\nabla_{\theta} \frac{1}{L} \sum_{j=1}^L f_w(g_{\theta}(z^{(j)})), \theta, \alpha, \beta_1, \beta_2)
    \]
\EndWhile
\end{algorithmic}
\end{algorithm}

Here, sample sequences are produced from the current generator $g_{\theta}(\cdot)$, while the discriminator/critic $f_{w}(\cdot)$ trains itself to distinguish between real and generated data by minimizing the Wasserstein loss with a gradient penalty. We use ADAM, a first-order gradient-based method widely used in the deep learning literature for optimization updates (Refer to \cite{ADAMoriginalpaper} for details). In each epoch, for $n_{\text{critic}}$ many ADAM updates to the discriminator, the generator is updated once. This process is continued until the empirical loss function $\hat{l}$ converges, leading to realistic data generation. 

\section{WGAN-based parametric estimation in Hawkes Process with missingness}
\label{sec:WGAN based estimate}
We recall that in many real-life scenarios, Hawkes process event streams often contain missing events due to non-reporting. This makes the likelihood intractable, as the exact positions and number of missing events are unknown. When a fraction ($p \in (0,1)$) of events is missing, estimating parameters $(\theta = (\mu, \alpha, \beta, \sigma^2)$ using maximum likelihood becomes impractical. The WGAN-based likelihood-free method for parameter estimation using exact generators affords a way out of this quandary.
\\

There are several methods to generate self-exciting spatiotemporal Hawkes processes in the existing literature, such as thinning, cluster-based-generation, etc. (\cite{reinhart2018review}). We use the method described in Section 5.1 of  \cite{ogata1988hawkesprocess}. This is specifically advantageous for our use case from computational and implementation standpoints since:
\begin{itemize}
    \item The expressions for time-stamps and locations of the events can be expressed as clear and concise algebraic expressions,  which is essential for deep learning frameworks, such as PyTorch or Tensorflow, to create a computational graph. 

    \item The calculations involved can be easily vectorized to generate batches of spatiotemporal Hawkes streams simultaneously, making it efficient to generate multiple synthetic data streams from the generator in every epoch while training the WGAN model (Refer to \ref{alg:our_algo_flowchart} for details). 
\end{itemize}

In our experiments, we generate \( K = 10,000 \) independent streams of Hawkes processes, denoted as \( \{s_1, s_2, \dots, s_K\} \), and randomly remove a percentage of events to simulate missing data.  The resultant streams, \( \{s_1^*, s_2^*, \dots, s_K^*\} \), may have different lengths due to the missing events. To maintain consistency, we pad the shorter streams with zeros to match the length of the longest stream. In the discriminator architecture, the weights corresponding to these padded zeros are kept inactive, a technique referred to as \textit{padding and masking} in deep learning (see \cite{paddinginLSTM}). This ensures that the model learns only from actual event data while ignoring the padding, allowing it to handle varying stream lengths effectively.
\\

We use a Wasserstein GAN (WGAN) model to estimate the parameters of the Hawkes process. The generator \(G(\cdot) \) is an exact simulator that generates a Hawkes process with parameters $\theta =(\mu, \alpha,\beta, \sigma^2)$ followed by introducing missingness at random (MAR) according to a given mechanism $p(x,y)$ in the streams, while the discriminator \( D \) is an LSTM network with a specified number of hidden nodes. Both the generator and discriminator are trained using the ADAM optimizer.

\subsection{Experiment using different initialization points}

Since the optimization problem in our method is non-convex, we usually start training the WGAN from many different initialization values of $\theta = (\mu, \alpha, \beta, \sigma^2)$. This is a standard tool for solving non-convex optimization problems.
After obtaining the parameter estimates from the WGAN for each initialization point, we evaluate the goodness of fit of these estimates $\hat\theta$ using novel criteria described below.

\subsubsection{ Goodness of Fit tests for Hawkes data with missingness}

The standard goodness of fit test for point processes is based on the compensator given by
$\Lambda(t)=\int_0^t \lambda(u) du$. Given a realization $\{t_1,t_2, \cdots, t_n\}$ of a temporal point process, we look at the values 
$$
\{\Lambda(t_1), \Lambda(t_2)-\Lambda(t_1), \cdots, \Lambda(t_n)-\Lambda(t_{n-1)}\}
$$

and check whether they follow an $\exp(1)$ distribution by making a QQ plot. See \cite{ogata1988hawkesprocess} and \cite{reinhart2018review} for the details. However, as we cannot explicitly write the intensity $\lambda(t,x,y)$ of the Spatiotemporal Hawkes model in the presence of missingness, this rules out the classical compensator-based goodness of fit test by QQ plotting. To the best of our knowledge, the current literature does not directly address this question. We provide below an empirical approach to the problem. 

Recall that the training data consists of $K$ many IID streams of reported crime data, each of size $N$. We propose statistics that are calculable from a given stream of data, which are then used to devise a goodness-of-fit criterion. \cite{deutsch2020abcHawkes} proposed some such statistics for their work on analyzing \textit{temporal} Hawkes processes with distortion. 
\\

Recall that the intensity of the \textit{complete} spatiotemporal Hawkes process is given by: 

$$
\lambda(t, x, y \mid \mathcal{H}_t)=\mu(x, y)+\sum_{t<t_i} g\left(t-t_i, x-x_i, y-y_i\right)
\,.$$

Further, assume that the triggering kernel has the product form 
$$
g(t, x, y)=g_1(t) g_2(x, y)
$$

where, $\int_0^{\infty} g_1(t) d t=A$ and $\int_{-\infty}^{\infty} \int_{-\infty}^{\infty} g(x, y) d x d y=\mathbb{B}$.
\\

Integrating the spatial coordinates $x$ and $y$ out, the intensity of the temporal process is given by:
$$
\begin{aligned}
& \lambda\left(t \mid H_t\right)=\int_{-\infty}^{\infty} \int_{-\infty}^{\infty} \lambda(t, x, y) d x d y \\
&=\iint_{\mathbb{R}^2}\mu(x,y) dxdy+\sum_{t_i<t} g_1\left(t-t_i\right) \iint_{\mathbb{R}^2} g_2\left(x-x_i ,y-y\right) d x d y\\
& = \mu + \sum_{t_i<t} B g_1\left(t-t_i\right).\\
\end{aligned}
$$

So, the temporal projection is a temporal Hawkes process with mean $\mu =\iint_{\mathbb{R}^2}\mu(x,y) $ and triggering kernel  $Bg_1(\cdot)$. Note that for our Bogota example, 
$$
B = \iint_{\mathbb{R}^2} \exp\{-\frac{1}{2\pi \sigma^2}(x^2+y^2) \} = 2\pi \sigma^2.
$$

So, while the interarrival times $\Delta_i=t_i-t_{i-1}$ do not explicitly depend on the location of the events, they do contain valuable information about the spatial dispersion parameter $\sigma^2$. However, we do not have the complete event series as we observe only reported events. We denote such event streams (involving missingness) as  $\{(t_i^*,x^*_i,y^*_i)\}_{i=1}^{N}$.
\vspace{0.1cm}
\\

The definitions below are based on a single stream $\{(t_i^*,x^*_i,y^*_i)\}_{i=1}^{N}$ and can be aggregated as needed. Define the inter-arrival times (from observed events) as:
$\Delta^*_i=t^*_i-t^*_{i-1}$ for a given stream of events. The goodness of fit is assessed by generating synthetic data streams based on the estimated parameter $\hat\theta$ and comparing the inter-arrival times observed in these synthetic data streams to those of the training data. For the comparison of the distribution of $\{\Delta_i^*\ \mid i \geq 1\}$ for observed data and simulated data, we use a Chi-Squared statistic (\cite{chisquaretestforhistograms}). We present our approach below.

\subsubsection*{Goodness of fit criterion}

\noindent\rule{\linewidth}{0.4pt}

\textbf{Input}: Reported crime data consisting of IID streams and an estimate $\hat{\theta} = (\hat{\mu}, \hat{\alpha}, \hat{\beta}, \hat{\sigma})$.

\noindent\rule{\linewidth}{0.4pt}

\begin{enumerate}
    \item Given an estimate $\hat{\theta}$, we generate $K_{synthetic}$ many (a large number) IID streams of spatiotemporal Hawkes process data from $P_{\hat{\theta}}$ followed by incorporating the appropriate missingness mechanism. 

    \item For the $j^{th}$ of the generated streams, calculate the set of interarrival times 
    $$
 I_j = \left\{\Delta_1^{(j)*}, \Delta_2^{(j)*}, \ldots, \ldots, \Delta_N^{(j)*} \right\}
    $$

    \textit{Note that the quantity $N$ might vary according to the stream owing to different lengths.}

    \item Calculate the union of all (generated) inter-arrival  times given by
$$
I_{\text{synthetic}}=\bigcup_{j=1}^{K_{\text {synthetic }}} I_j
$$

 \item Using the training data, calculate an analogous set of all interarrival times from all the streams. Call it $I_{\text{training}}$.

    \item 
    We break the range (union of the supports of $I_{training}$ and $I_{synthetic}$ ) into bins of equal-width intervals and compare the dissimilarity using the chi-square statistic:

$$
\chi^2 = \sum_{i=1}^{n_{\text{bins}}} \frac{(f^i_{\text{training}}-f^i_{\text{synthetic}})^2}{f^i_{\text{training}}}
$$

where $f^i_{\text{training}}$ and $f^i_{\text{synthetic}}$ denote the frequency of the $i^{th}$ bin for the histograms  of $I_{\text{training}}$  and $I_{\text{synthetic}}$, respectively.
    
\end{enumerate}

\noindent\rule{\linewidth}{0.4pt}

Given the IID streams of training samples, we can run our proposed estimation method starting from several initialization points for $\theta$ (say, $\theta_{init}$) and choose the best estimate (i.e., the minimum value of the $\chi^2$ statistics) according to the goodness of fit criterion described above. 
    
\end{doublespace}
\begin{doublespace}

\section{WGAN-based learning for Bogota data accounting for missingness}

\noindent \textbf{Description of Real data:} 

\begin{itemize}
    \item We first simulate $10,000$ IID realizations of crime data series from the spatiotemporal process described below:

\begin{equation}
\mu(t, x, y)=\mu \sum_{c_i} \frac{1}{2 \pi \sigma_0^2} \exp \left(-\frac{\left\|(x, y)-c_i\right\|^2}{2 \sigma_0^2}\right)
\end{equation}

where $c_i$ 's are $14$ centers across the Bogota grid map, namely 
$$( \pm 6, \pm 20),( \pm 6, \pm 10),( \pm 6, \pm 30),( \pm 6,0)
$$ and $\sigma_0^2=(4.5)^2$. The triggering kernel is a product kernel with exponential decay in the temporal dimension and a Gaussian kernel in the spatial component given by 

\begin{equation}
g(t, x, y)=\alpha\exp (-\beta t) \exp \left(-\frac{x^2+y^2}{2 \sigma^2}\right)\,.
\end{equation}

We generate the first $250$ events in each stream.

\item  Followed by the full generation above, we randomly remove crimes in each stream according to district-wise missingness rates ( = $1$ -- reporting rates) described \cite{predictivepolicing} (see Figure \ref{fig:Bogota map}). Note that, for any location $(x,y) \in \text{ district } \: d$, the distortion function $p(x,y)$, referred to in section \ref{sec:Our method}, is equal to the missingness rate in district $d$. This is an example of a variable missingness rate where the chance of observing an event depends on the location. 
\end{itemize}

\noindent \textit{Training the WGAN model:}
The WGAN model is trained using an exact spatiotemporal Hawkes process generator with district-specific thinning. The discriminator is an LSTM with 64 hidden layer nodes. Both the generator and discriminator are updated using the ADAM optimizer with a batch size of $L=256$. We also start with several different initialization points for $\theta$ over a grid and choose the best estimate using the goodness of fit criterion described above. 
\newline
\newline
\textbf{Hotspot prediction:} Recall that the quality of recovery of Hawkes process parameters when training the EM algorithm only on reported crime data (see Table \ref{tab:Bogota_EM_estimates_table} in Section  \ref{sec:Intro to predictive policing in Bogota}) is poor, and therefore compromises the quality of downstream tasks, like hotspot predictions. Below, we demonstrate that our proposed approach reliably estimates the parameters of the spatiotemporal process even in the presence of substantial missingness rates, which leads to quality hotspot prediction. 
\\

First, for ease of computation, we divide the rectangular grid map in Figure 12 into a $7 \times 16$ grid, i.e., we divide the $x$ range into $7$ equal parts and the $y$ range into $16$ equal parts.  For each of $P_{\theta_0}$ and $P_{\hat{\theta}}$, we simulate $100$ independent Monte Carlo samples of crime streams in Bogota, followed by thinning according to district-specific reporting rates, over a week, i.e., $t \in [0,T]$ for $T=7$. 



\begin{itemize}
    \item We compute $E_{\theta}[N_A]$:  the expected average number of daily reported crimes in grid cell $A$ over  the week for $\theta\in \{\theta_0, \hat \theta\}$ by generating Monte Carlo samples (followed by thinning) and then averaging cell counts from each realization.

    \item To assess the reliability of our estimates, we compare $E_{\theta_0}[ N_{A}]$ (true mean) and $E_{\hat\theta}[ N_{A}]$ (estimated mean) using the relative mean absolute error (MAE) :

 $$
\frac{\left|E_{\theta_0}\left(N_{ A}\right)-E_{\hat{\theta}}\left(N_{ A}\right)\right|}{E_{\theta_0}\left(N_{ A}\right)} \,.
$$
\end{itemize}

We find the top $10$ cells corresponding to the highest values of $E_{\theta}(N_A)$ , referring to these cells as \textit{hotspots}, for   $\theta \in \{\theta_0, \hat\theta\}$, and 
check the percentage of common hotspots between $\theta_0$ and $\hat \theta$. We call this \textit{top $10$ accuracy} for reported crime hotspot prediction using our estimated parameters.

\begin{figure}[h!]
    \centering
    \begin{subfigure}[b]{0.4\textwidth}
        \centering
        \includegraphics[width=\textwidth]{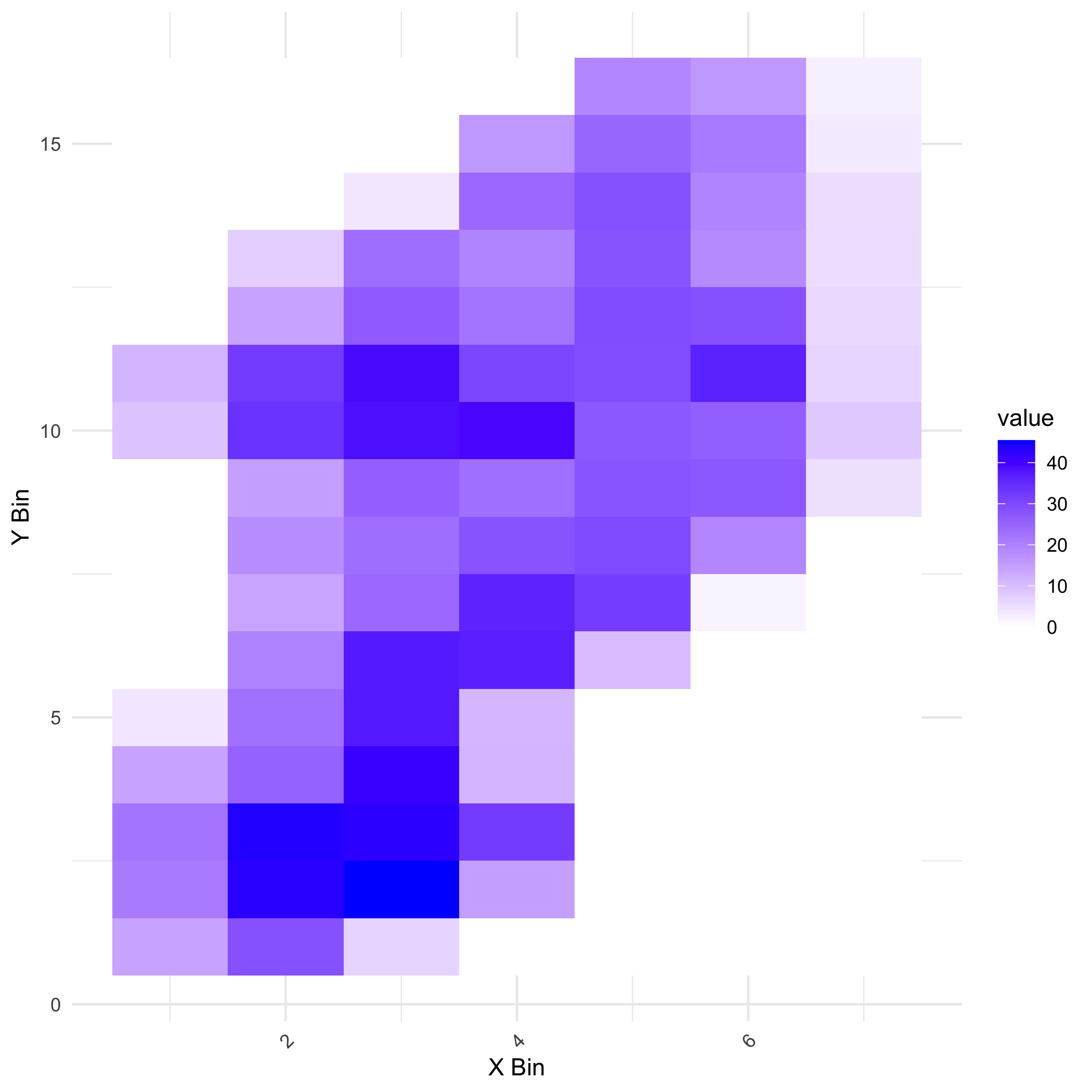}
        \caption{True (w.r.t. \(P_{\theta_0}\)) expectation of average daily crime counts in the grid}
        \label{fig:true_heatmap}
    \end{subfigure}
    \hfill
    \begin{subfigure}[b]{0.4\textwidth}
        \centering
        \includegraphics[width=\textwidth]{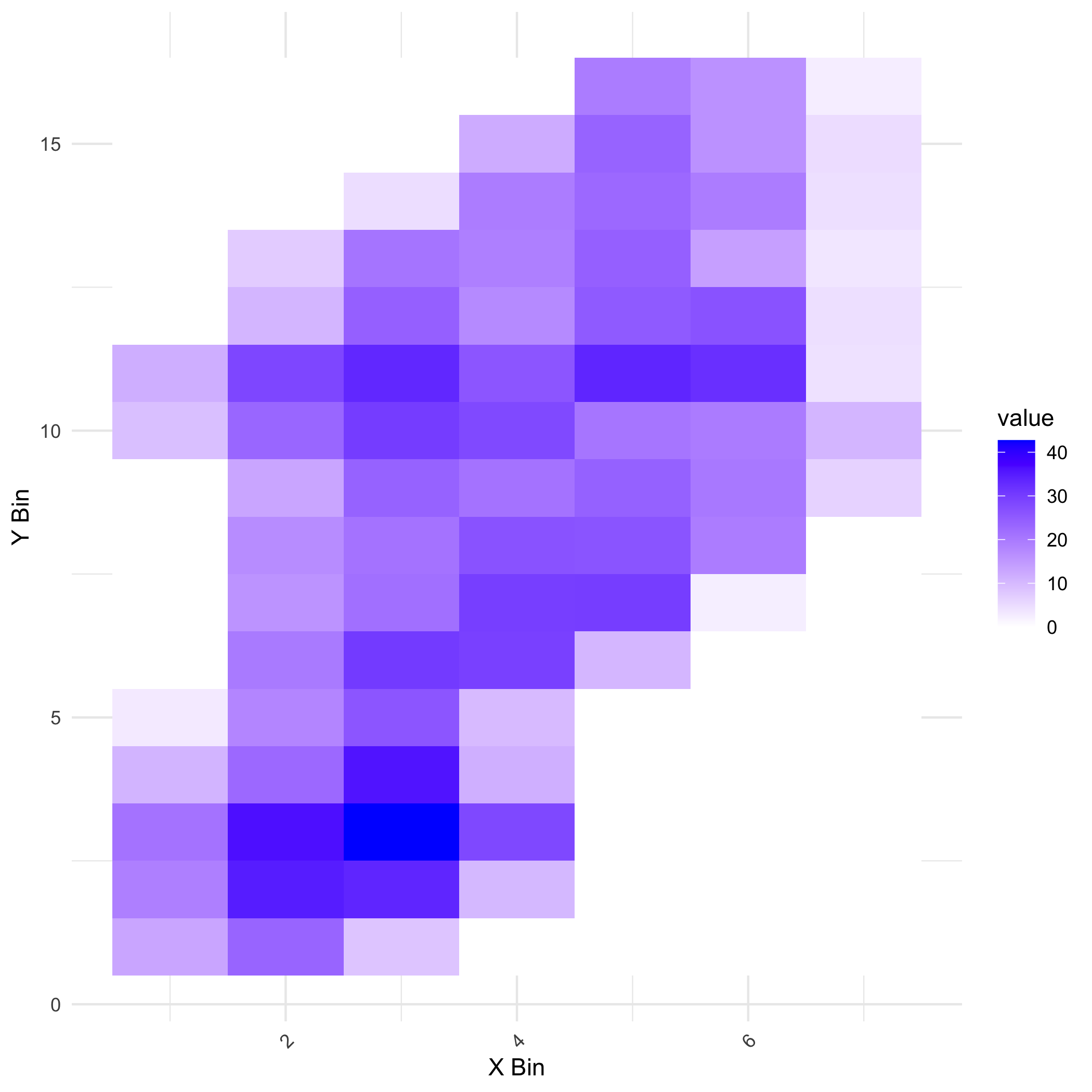}
        \caption{Estimated (w.r.t. \(P_{\hat{\theta}}\)) expectation of average daily crime counts in the grid}
        \label{fig:estimated_heatmap}
    \end{subfigure}
    \caption{Comparison of true vs. estimated crime hotspots for the Bogota crime data}
    \label{fig:heatmap_comparison}
\end{figure}

\vspace{0.2cm}

The $P_{\hat\theta}$ hotspots are given by 
$$
\{(2,3),(3,2),(3,11),(2,2),(3,3),(6,11),(3,6),(3,4), (5,11), (3,10)\}
$$
while the $P_{\theta_0}$ hotspots are 
$$
\{(3,3),(2,3),(3,2),(3,4),(2,2),(3,5),(3,6),((3,10),(3,11),(4,10)\} \,.
$$

Note that these $2$ sets share 8  elements out of $10$, i.e. the top-$10$ accuracy is $80 \%$.

The MAE, averaged over the $100$ Monte Carlo replicates and all cells, is only $10.219 \%$. This shows that we can identify the hotspots and anticipated reported crime counts in each area with reasonable accuracy, even when training the model on data with significant missingness.  
\newline
\newline
{\bf Robustness of the Hotspot prediction results:}
The choice of the true Hawkes process parameters used in \cite{NiljanaAkpinarpredpolicing_2021} for simulating the crime data in Bogota is 
$$
\theta_0=(\mu,\alpha,\beta, \sigma^2)=(100,3,0.2,0.1^2).
$$ 
However, this choice does not appear to be explicitly linked to existing domain knowledge or results from prior studies. Hence, we test the robustness of our method by running it on crime data generated from \textit{true parameters} given by

$$
\theta_0=(\mu,\alpha,\beta, \sigma^2) \in \{95,100,105\}\times \{2,3,4\}\times \{0.1,0.2,0.3\} \times \{0.1^2\} \,,
$$

and repeat the hotspot comparison exercise described above.

We find that the top $10$ accuracy figures for predicting reported crime hotspots for the 27 parameter combinations vary between $60$ and $90 \%$ with an average of $74.8\%$.  In other words, we can recover around $7$ out-of-the-top $10$ hotspots even with significant under-reporting.
\end{doublespace}

\begin{doublespace}

\section{Discussion}

In this paper,  we propose a novel method for likelihood-free estimation of Spatiotemporal Hawkes processes with missing events. This is a particularly challenging problem in predictive policing, where only recorded crimes serve as training data, rendering maximum likelihood estimation infeasible due to the intractability of the resulting likelihood function.  This framework was originally examined in \cite{NiljanaAkpinarpredpolicing_2021}, which highlighted the challenges of crime under-reporting and its effects on subsequent tasks in predictive policing.

To address this problem, we leverage Wasserstein Generative Adversarial Networks (WGANs) (\cite{arjovsky2017wasserstein}) to estimate the Hawkes process parameters directly from the reported crime data, even when a significant portion of the data is unreported. The WGAN-based approach is particularly suited for this problem because it operates within a likelihood-free framework, making it robust to the challenges posed by missing data. Unlike traditional methods, our approach does not require explicit knowledge of the missing events' positions or numbers, which are unavailable in practice. Our results demonstrate that the WGAN-based method reliably recovers the underlying parameters of the Hawkes process, even when trained on data with substantial missingness.

By incorporating differential reporting rates across various spatial regions into our estimation method, we contribute to developing more accountable predictive policing algorithms in contrast to traditional methods that often overlook these variations, leading to biased predictions that result in over-policing or under-policing, with harmful implications for vulnerable communities (\cite{NiljanaAkpinarpredpolicing_2021}). 

From an applied standpoint, a natural extension of our work is to investigate cases where the missingness rate is not known beforehand, in which case identifiability of the Hawkes process parameters is not necessarily guaranteed. To this end, we include a lemma establishing identifiability of these parameters under a natural constraint on the reporting function in Supplement \ref{sec: identifiability}. Further research into these issues would significantly broaden the applicability of our approach to real-world settings where reporting rates are often difficult to ascertain and may vary over time and space.

While Hawkes processes provide a powerful framework for modeling spatiotemporal events exhibiting self-exciting behavior, such as crime data (\cite{UCLAcrimepaper}), the advent of deep learning-based generators, such as Convolutional Long Short-term memory networks (ConvLSTM) and self-attention-based models (\cite{vaswani2017attentionisallyouneed}), offer the potential to capture even richer patterns in the spatiotemporal dynamics (\cite{YaoXiecrime911calltextAtlanta2020spatial},\cite{DivyaSaxenaConvLSTMDCGAN}, \cite{mei2017neuralhawkesprocess}). In the future, we plan to extend our methods by using flexible model-agnostic spatiotemporal  generators for modeling reported crime data in large cities, such as Los Angeles and Chicago. 
\newline
\newline
In conclusion, our work represents a significant step towards developing robust, accountable, and transparent predictive policing tools that can operate effectively despite incomplete data. By addressing the challenges of crime underreporting and adopting generative machine learning in this domain, we lay the fundamentals for future development that will continue refining predictive policing models' capabilities in public safety applications. 
\end{doublespace}

\bibliography{sm.bib}

\begin{thebibliography}{29}
\providecommand{\natexlab}[1]{#1}
\providecommand{\url}[1]{\texttt{#1}}
\expandafter\ifx\csname urlstyle\endcsname\relax
  \providecommand{\doi}[1]{doi: #1}\else
  \providecommand{\doi}{doi: \begingroup \urlstyle{rm}\Url}\fi

\bibitem[Achab et~al.(2017)Achab, Bacry, Ga{\i}ffas, Mastromatteo, and Muzy]{AchabNHPCgrangercausalitycummulants}
Massil Achab, Emmanuel Bacry, St{\'e}phane Ga{\i}ffas, Iacopo Mastromatteo, and Jean-Fran{\c{c}}ois Muzy.
\newblock Uncovering causality from multivariate hawkes integrated cumulants.
\newblock In \emph{International Conference on Machine Learning}, pages 1--10. PMLR, 2017.

\bibitem[Akpinar et~al.(2021{\natexlab{a}})Akpinar, De-Arteaga, and Chouldechova]{NiljanaAkpinarpredpolicing_2021}
Nil-Jana Akpinar, Maria De-Arteaga, and Alexandra Chouldechova.
\newblock The effect of differential victim crime reporting on predictive policing systems.
\newblock In \emph{Proceedings of the 2021 {ACM} Conference on Fairness, Accountability, and Transparency}. {ACM}, mar 2021{\natexlab{a}}.
\newblock \doi{10.1145/3442188.3445877}.
\newblock URL \url{https://doi.org/10.1145%2F3442188.3445877}.

\bibitem[Akpinar et~al.(2021{\natexlab{b}})Akpinar, De{-}Arteaga, and Chouldechova]{predictivepolicing}
Nil{-}Jana Akpinar, Maria De{-}Arteaga, and Alexandra Chouldechova.
\newblock The effect of differential victim crime reporting on predictive policing systems.
\newblock \emph{CoRR}, abs/2102.00128, 2021{\natexlab{b}}.
\newblock URL \url{https://arxiv.org/abs/2102.00128}.

\bibitem[Arjovsky et~al.(2017)Arjovsky, Chintala, and Bottou]{arjovsky2017wasserstein}
Martin Arjovsky, Soumith Chintala, and L{\'e}on Bottou.
\newblock Wasserstein generative adversarial networks.
\newblock In \emph{International conference on machine learning}, pages 214--223. PMLR, 2017.

\bibitem[Bacry et~al.(2015)Bacry, Mastromatteo, and Muzy]{bacry2015hawkes}
Emmanuel Bacry, Iacopo Mastromatteo, and Jean-Fran{\c{c}}ois Muzy.
\newblock Hawkes processes in finance.
\newblock \emph{Market Microstructure and Liquidity}, 1\penalty0 (01):\penalty0 1550005, 2015.

\bibitem[Deutsch and Ross(2020)]{deutsch2020abcHawkes}
Isabella Deutsch and Gordon~J Ross.
\newblock Abc learning of hawkes processes with missing or noisy event times.
\newblock \emph{arXiv preprint arXiv:2006.09015}, 2020.

\bibitem[Drossos and Philippou(1980)]{MDE_note}
Constantine~A Drossos and Andreas~N Philippou.
\newblock A note on minimum distance estimates.
\newblock \emph{Annals of the Institute of Statistical Mathematics}, 32:\penalty0 121--123, 1980.

\bibitem[Dwarampudi and Reddy(2019)]{paddinginLSTM}
Mahidhar Dwarampudi and NV~Reddy.
\newblock Effects of padding on lstms and cnns.
\newblock \emph{arXiv preprint arXiv:1903.07288}, 2019.

\bibitem[Gagunashvili(2006)]{chisquaretestforhistograms}
ND~Gagunashvili.
\newblock $\chi$2 test for the comparison of weighted and unweighted histograms.
\newblock In \emph{Statistical Problems In Particle Physics, Astrophysics And Cosmology}, pages 43--44. World Scientific, 2006.

\bibitem[Goodfellow et~al.(2014)Goodfellow, Pouget-Abadie, Mirza, Xu, Warde-Farley, Ozair, Courville, and Bengio]{GANoriginalpaper}
Ian Goodfellow, Jean Pouget-Abadie, Mehdi Mirza, Bing Xu, David Warde-Farley, Sherjil Ozair, Aaron Courville, and Yoshua Bengio.
\newblock Generative adversarial nets.
\newblock \emph{Advances in neural information processing systems}, 27, 2014.

\bibitem[Gulrajani et~al.(2017)Gulrajani, Ahmed, Arjovsky, Dumoulin, and Courville]{Gradientpenalty}
Ishaan Gulrajani, Faruk Ahmed, Martin Arjovsky, Vincent Dumoulin, and Aaron~C Courville.
\newblock Improved training of wasserstein gans.
\newblock \emph{Advances in neural information processing systems}, 30, 2017.

\bibitem[Hawkes(2018)]{hawkes2018hawkesfinance}
Alan~G Hawkes.
\newblock Hawkes processes and their applications to finance: a review.
\newblock \emph{Quantitative Finance}, 18\penalty0 (2):\penalty0 193--198, 2018.

\bibitem[Hawkes and Oakes(1974)]{hawkes1974cluster}
Alan~G Hawkes and David Oakes.
\newblock A cluster process representation of a self-exciting process.
\newblock \emph{Journal of applied probability}, 11\penalty0 (3):\penalty0 493--503, 1974.

\bibitem[Hochreiter and Schmidhuber(1997)]{hochreiter1997originalLSTM}
Sepp Hochreiter and J{\"u}rgen Schmidhuber.
\newblock Long short-term memory.
\newblock \emph{Neural computation}, 9\penalty0 (8):\penalty0 1735--1780, 1997.

\bibitem[Kingma(2014)]{ADAMoriginalpaper}
Diederik~P Kingma.
\newblock Adam: A method for stochastic optimization.
\newblock \emph{arXiv preprint arXiv:1412.6980}, 2014.

\bibitem[Mei and Eisner(2017)]{mei2017neuralhawkesprocess}
Hongyuan Mei and Jason~M Eisner.
\newblock The neural hawkes process: A neurally self-modulating multivariate point process.
\newblock \emph{Advances in neural information processing systems}, 30, 2017.

\bibitem[Mohler et~al.(2011)Mohler, Short, Brantingham, Schoenberg, and Tita]{UCLAcrimepaper}
George~O Mohler, Martin~B Short, P~Jeffrey Brantingham, Frederic~Paik Schoenberg, and George~E Tita.
\newblock Self-exciting point process modeling of crime.
\newblock \emph{Journal of the American Statistical Association}, 106\penalty0 (493):\penalty0 100--108, 2011.

\bibitem[Ogata(1988)]{ogata1988hawkesprocess}
Yosihiko Ogata.
\newblock Statistical models for earthquake occurrences and residual analysis for point processes.
\newblock \emph{Journal of the American Statistical association}, 83\penalty0 (401):\penalty0 9--27, 1988.

\bibitem[Olaoye and Egon(2024)]{favour2024predictivepolicing}
Favour Olaoye and Axel Egon.
\newblock Predictive policing and crime prevention.
\newblock \emph{Crime \& Delinquency}, 08 2024.

\bibitem[Reinhart(2018)]{reinhart2018review}
Alex Reinhart.
\newblock A review of self-exciting spatio-temporal point processes and their applications.
\newblock \emph{Statistical Science}, 33\penalty0 (3):\penalty0 299--318, 2018.

\bibitem[Rossant et~al.(2011)Rossant, Leijon, Magnusson, and Brette]{neurosciencerossant2011sensitivity}
Cyrille Rossant, Sara Leijon, Anna~K Magnusson, and Romain Brette.
\newblock Sensitivity of noisy neurons to coincident inputs.
\newblock \emph{Journal of neuroscience}, 31\penalty0 (47):\penalty0 17193--17206, 2011.

\bibitem[Saxena and Cao(2019)]{DivyaSaxenaConvLSTMDCGAN}
Divya Saxena and Jiannong Cao.
\newblock D-gan: Deep generative adversarial nets for spatio-temporal prediction.
\newblock \emph{arXiv preprint arXiv:1907.08556}, 2019.

\bibitem[Saxena and Cao(2021)]{ChallengesoftrainingGANDivyaSaxena}
Divya Saxena and Jiannong Cao.
\newblock Generative adversarial networks (gans) challenges, solutions, and future directions.
\newblock \emph{ACM Computing Surveys (CSUR)}, 54\penalty0 (3):\penalty0 1--42, 2021.

\bibitem[Staudemeyer and Morris(2019)]{LSTMreview}
Ralf~C Staudemeyer and Eric~Rothstein Morris.
\newblock Understanding lstm--a tutorial into long short-term memory recurrent neural networks.
\newblock \emph{arXiv preprint arXiv:1909.09586}, 2019.

\bibitem[Vaswani(2017)]{vaswani2017attentionisallyouneed}
A~Vaswani.
\newblock Attention is all you need.
\newblock \emph{Advances in Neural Information Processing Systems}, 2017.

\bibitem[Veen and Schoenberg(2008)]{JASAearthquake}
Alejandro Veen and Frederic~P Schoenberg.
\newblock Estimation of space–time branching process models in seismology using an em–type algorithm.
\newblock \emph{Journal of the American Statistical Association}, 103\penalty0 (482):\penalty0 614--624, 2008.
\newblock \doi{10.1198/016214508000000148}.
\newblock URL \url{https://doi.org/10.1198/016214508000000148}.

\bibitem[Xiao et~al.(2017)Xiao, Farajtabar, Ye, Yan, Song, and Zha]{WGAN+TPP}
Shuai Xiao, Mehrdad Farajtabar, Xiaojing Ye, Junchi Yan, Le~Song, and Hongyuan Zha.
\newblock Wasserstein learning of deep generative point process models, 2017.
\newblock URL \url{https://arxiv.org/abs/1705.08051}.

\bibitem[Yuan et~al.(2019)Yuan, Li, Bertozzi, Brantingham, and Porter]{networkreconstructionSThawkes}
Baichuan Yuan, Hao Li, Andrea~L Bertozzi, P~Jeffrey Brantingham, and Mason~A Porter.
\newblock Multivariate spatiotemporal hawkes processes and network reconstruction.
\newblock \emph{SIAM Journal on Mathematics of Data Science}, 1\penalty0 (2):\penalty0 356--382, 2019.

\bibitem[Zhu and Xie(2021)]{YaoXiecrime911calltextAtlanta2020spatial}
Shixiang Zhu and Yao Xie.
\newblock Spatial-temporal-textual point processes for crime linkage detection, 2021.
\newblock URL \url{https://arxiv.org/abs/1902.00440}.

\end{thebibliography}

\begin{doublespace}

\section{Supplementary Materials}

\subsection{Branching structure and EM algorithm for inference in Hawkes process}
\label{sec:Hawkes EM algorithm}

In a self-exciting Hawkes process, an event is either generated from the background intensity $\mu_0(\cdot)$, or is triggered by a past event via the triggering intensities. Consequently, each event can be classified it into one of two types (\cite{hawkes1974cluster},\cite{AchabNHPCgrangercausalitycummulants}):
\begin{itemize}
    \item Immigrants: events generated from the background process
    \item Offspring: events generated from a past parent event via triggering.
\end{itemize}

It is, therefore, possible to represent the Hawkes process as a branching process where immigrant events are the roots of trees and which, eventually, have offspring down the tree. 
The branching representation allows the use of the EM algorithm for the estimation of the Hawkes process parameters $\theta = (\mu_0, \alpha, \beta, \sigma^2)$.  With the Hawkes process, we observe just  the events (time stamps and (or) spatial location marks),  but \emph{not} the ancestral history of any given event, which is contained in the (unobserved) branching structure. Hence, the Hawkes process data is viewed as a missing data problem,  while the full data, which incorporates the branching structure of the events, has a simpler  likelihood structure, exploitable by the EM algorithm.  For example, \cite{JASAearthquake} maximize the log-likelihood  via EM by using latent variables $u_i$ for each event $i$,  defined as 

 $$
u_i = \begin{cases}
         j & \text{if } j<i \text{ and event $j$ triggers event $i$} \\
         i & \text{if }\text{ event $i$ is a backgroud event}\,.
       \end{cases}
$$

Assuming knowledge of the branching structure i.e., $u_i$'s, the complete-data loglikelihood for $\theta$ is given by
\begin{small}

\begin{eqnarray*}
l(\theta, \{x_i, y_i, u_i\}) &=& \sum_{i=1}^n \mathbb{I}\left(u_i=i\right) \log \left(\mu\left(x_i,y_i\right)\right)\\
& + & \sum_{i=1}^n \sum_{j=1}^n \mathbb{I}\left(u_i=j\right) \log \left(g\left(t_i-t_j,x_i-x_j, y_i-y_j\right)\right)\\
&-& \int_0^T \int_X \lambda(t,x,y \mid \theta) \mathrm{~d}t\,\mathrm{d}x\,\mathrm{d}y \,.
\end{eqnarray*}
\end{small}

\begin{itemize}
\item 
We start with some initial value $\theta_0 $. Subsequently, in each iteration, for the \textbf{E step}, the algorithm computes $E_{\theta_k}[l(\theta, \{x_i, y_i, u_i\})|\{x_i, y_i\})]$, where $\theta_k$ is the parameter value at the current iteration. The key formulas in this step are the conditional expectations of the indicator variables given by: 
$$
\begin{aligned}
& E_{\theta_k}[\mathbb{I}_{(u_i =j)}\mid \{x_i, y_i\}]
= \frac{g\left( t_i-t_j,x_i-x_j, y_i-y_j \mid \theta_k \right)}{\lambda\left( t_i,x_i,y_i \mid \theta_k \right)} \text{ if } j<i \,,\\
\\
& E_{\theta_k}[\mathbb{I}_{(u_i =i)}\mid \{x_i, y_i\}] 
=\frac{\mu\left(x_i,y_i \mid \theta_k \right)}{\lambda\left(t_i,x_i,y_i \mid \theta_k\right)} \,.
\end{aligned}
$$

\item In the \textbf{M-step}, the algorithm maximizes the conditional expectation in the above step to update $\theta$, which is done numerically in the absence of an exact analytical solution.   This iterative process is continued until convergence. 
\end{itemize}

\subsection{Some comments on Identifiability of the Hawkes process parameters}
\label{sec: identifiability}
While for the setup in this study, we assume that the exact reporting rates are known, often, this is not the case. Our result below shows that, under certain conditions, the reporting rate can be learned from the data. 
\newline

\textbf{Lemma:} Suppose $\lambda_i(t,x,y), i=1,2$ are Hawkes intensity functions with parameters $(\mu_i,\sigma_i),(\alpha_i,\beta_i, \sigma^{*2}_i)$ respectively, i.e.,

$$
\begin{aligned}
\mu_i(x,y) = \mu_ie^{-\frac{x^2+y^2}{2\sigma_i^2}}\\
g_i(t,x,y)= \alpha_i e^{-\beta_i t} e^{-\frac{x^2+y^2}{2 \sigma_1^{* 2}}}    
\end{aligned}
$$

and $h_1(x,y),h_2(x,y)$ are distortion functions taking positive values $\forall (x,y) \in \mathbf{R}^2$, such that 
$$
\begin{gathered}
\lambda_1\left(t, x, y \mid \mathcal{H}_t\right) h_1(x, y) =\lambda_2\left(t, x, y \mid \mathcal{H}_t\right) h_2(x, y) \:\:\forall (t, x, y) 
\end{gathered}
$$

Then, $\lambda_1(t,x,y\mid \mathcal{H}_t) \propto \lambda_2(t,x,y\mid \mathcal{H}_t )$.
\\

\textbf{Proof of the lemma:} Suppose the given sample path is $\{(t_i,x_i,y_i)\}_{i \geq 1}$.  Take $t <t_1$ yielding 

\begin{equation}
\begin{aligned}
\label{eqn: equation t<t_1}
& \mu_1 e^{-\frac{x^2+y^2}{2 \sigma_1^2}} h_1(x, y) =\mu_2 e^{-\frac{x^2+y^2}{2 \sigma_2^2}} h_2(x, y)\quad \forall x,y
\end{aligned}
\end{equation}

Now, take $t \downarrow t_1$ and 

\begin{equation}
\begin{aligned}
\label{eqn: t>t_1}
\alpha_1 e^{-\frac{(x-x_1)^2+(y-y_1)^2}{2 \sigma^{*2}_1}} h_1(x, y) =\alpha_2 e^{-\frac{(x-x_1)^2+(y-y_1)^2}{2 \sigma^{*2}_2}} h_2(x, y)\quad \forall x,y
\end{aligned}
\end{equation}

Then, by dividing these 2 equations, we get that

\begin{equation}
\label{eqn: equation}
\begin{gathered}
\frac{\alpha_1}{\mu_1} \cdot e^{-\frac{\left(x-x_1\right)^2+\left(y-y_1\right)^2}{2 \sigma_1^{* 2}}+\frac{x^2+y^2}{2 \sigma_1^2}} =\frac{\alpha_2}{\mu_2} \cdot e^{-\frac{\left(x-x_1\right)^2+(y-y_1)^2}{2 \sigma^*_2{ }^2}+\frac{x^2+y^2}{2 \sigma_2^2}} \forall x,y\\
\end{gathered}
\end{equation}

Since  \ref{eqn: equation}  is an identity in $(x,y)$, comparing the exponents on both sides, we get that $$
\begin{aligned}
& \sigma_1^2=\sigma_2^2 \\
& \sigma_1^{* 2}=\sigma_2^{* 2}\\
& \frac{\alpha_1}{\mu_1} = \frac{\alpha_2}{\mu_2}
\end{aligned}
$$

Now, taking $t \uparrow t_2$ and using \ref{eqn: equation t<t_1}, we get that,

$$
\begin{aligned}
& \alpha_1 e^{-\beta_1\left(t_2-t_1\right)} \cdot e^{-\frac{\left(x-x_1\right)^2+\left(y- y_1\right)^2}{2 \sigma_1^{* 2}}}  h_1(x, y)=\alpha_2 e^{-\beta_2\left(t_2-t_1\right)} e^{-\frac{\left(x-x_1\right)^2+\left(y- y_1\right)^2}{2 \sigma_2^{* 2}}}  h_2(x, y)
\end{aligned}
$$

Now, divide this by \ref{eqn: t>t_1} to get:

$$
e^{-\beta_!(t_2-t_1)} =e^{-\beta_2(t_2-t_1)}
$$

which shows that $\beta_1=\beta_2$.  \textbf{QED}.
\\

By the above lemma, even if we do not know the distortion function $h(\cdot)$, a scaling condition of the kind $\iint_{(x, y) \in S} h(x, y) d x d y=1$ identifies both $h(\cdot)$ and $\lambda(\cdot)$ uniquely. Thus, the parameters of the Hawkes process are identifiable under the reasonable assumption that the distortion function has a unit (or finite) integral over the spatial domain.

\subsection{Sequential deep learning models - Recurrent Neural nets and Long Short term memory network} 
\label{sec:LSTM+RNN_math}
Here, we discuss some of the mathematical formulations of sequential deep learning models. For translation, if the input sequence batch is $\zeta=\left\{x_1, \ldots, x_n\right\}$ and the output sequence batch is $\rho=\left\{t_1, \ldots, t_n\right\}$ then the generator $g_\theta(\zeta)=\rho$ transforms by:
$$
h_i=\phi_1\left(A_g^h x_i+B_g^h h_{i-1}+b_g^h\right), \quad t_i=\phi_2\left(B_g^x h_i+b_g^x\right)
$$

\begin{figure}[h!]
    \centering
    \includegraphics[width=6cm]{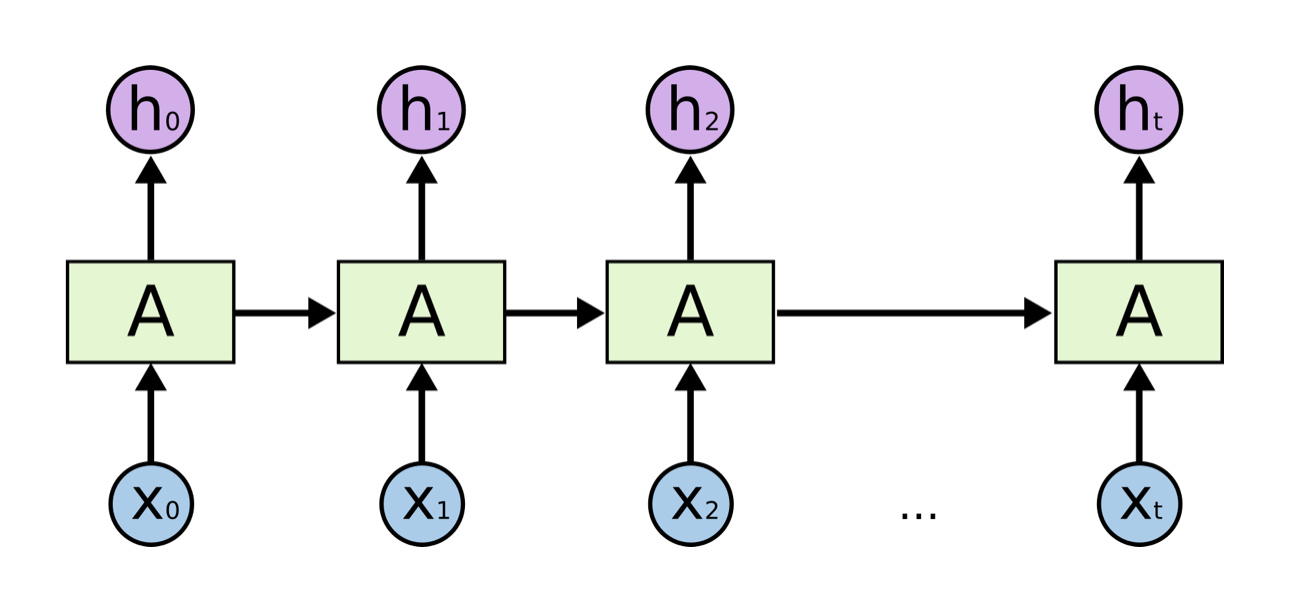}
    \caption{Architecture of RNN  from \href{https://colah.github.io/posts/2015-08-Understanding-LSTMs/}{Christopher Olah's blog}}
    \label{fig:RNN}
\end{figure}

\noindent where $h_i\in \mathbf{R}^k$ are history embedding vectors  (memory) and $\phi_1$ and $\phi_2$ are (vectorized) non-linear activation functions such as, ReLU, tanh, sigmoid, etc.. In this case, the generator is characterized by 
$$
\theta=\left\{\left(A_g^h\right)_{k \times 1},\left(B_g^h\right)_{k \times k},\left(b_g^h\right)_{k \times 1},\left(B_g^x\right)_{1 \times k},\left(b_g^x\right)_{1 \times 1}\right\}
$$. 

\noindent For using RNN's as a discriminator/critic, which gives a scalar value, we denote:
$$
f_w(\rho)=\sum_{i=1}^n\phi_d^a\left(B_d^a h_i+b_d^a\right) 
$$ to the input sequence $\rho=\left\{t_1, \ldots, t_n\right\}$, where the embedding (memory) vector is defined recursively as:
$$
h_i=\phi_d^h\left(A_d^h t_i+B_g^h h_{i-1}+b_g^h\right) \quad
$$

\noindent Here, the discriminator is  parameterized by the weights 
$$
w=\left\{\left(A_d^h\right)_{k \times 1},\left(B_d^h\right)_{k \times k},\left(b_d^h\right)_{k \times 1},\left(B_d^a\right)_{1 \times k},\left(b_d^a\right)_{1 \times 1}\right\}
$$

\begin{figure}[h!]

    \centering
    \includegraphics[width=10cm]{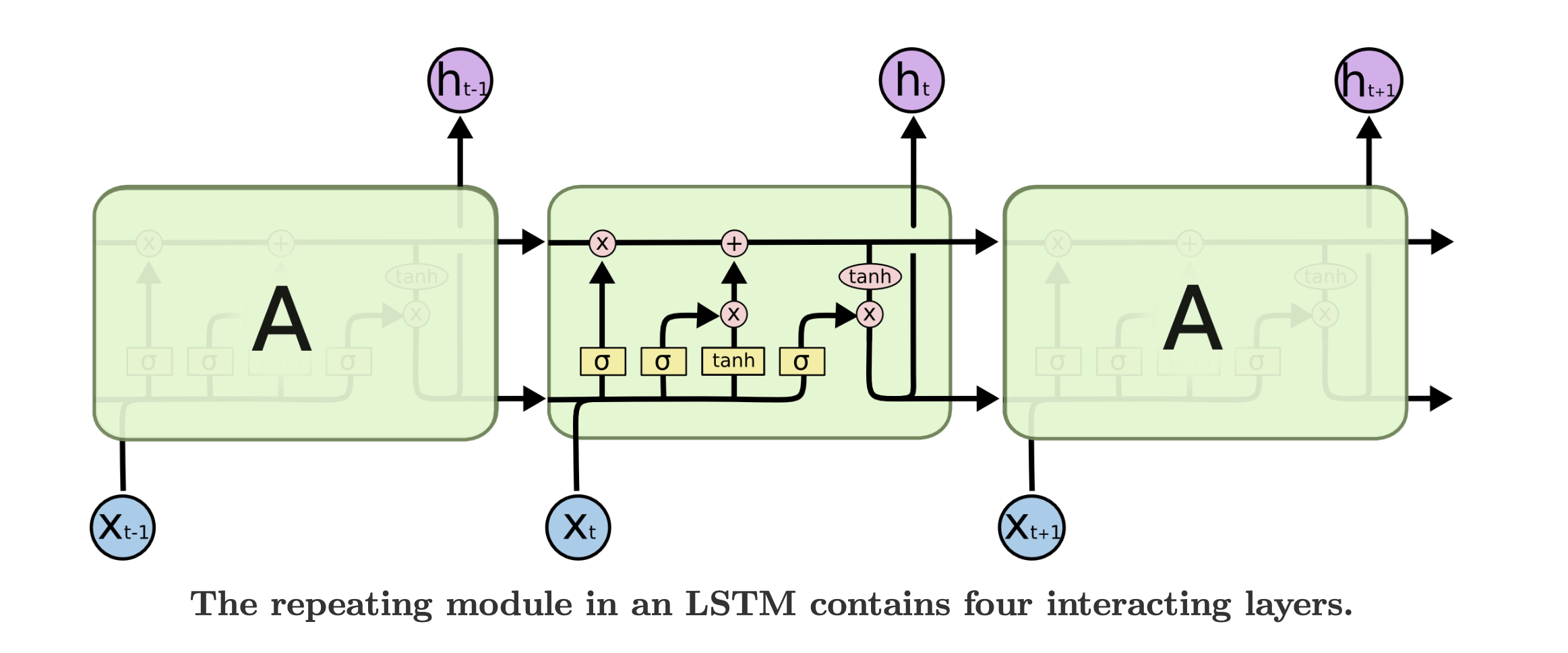}
    \caption{Structure of LSTM from \href{https://colah.github.io/posts/2015-08-Understanding-LSTMs/}{Christopher Olah's blog}}
    \label{fig:LSTM_gates}
    \label{fig:LSTM}
\end{figure}

To address common issues in training RNN, such as vanishing/exploding gradients as well as the inability to have a longer memory, \cite{hochreiter1997originalLSTM} introduced Long Short-Term Memory (LSTM) networks consisting of 3 different gates that control the information flow in the network (See Figure \ref{fig:LSTM_gates}). 

Combining history $h_{t-1}$ from time $t-1$ and new data $x_t$ at time $t$, the forget gate decides how much information will be thrown away by using a sigmoid layer. Next, the network computes what new information will be stored in the cell state. This is usually a two-step process - first, a sigmoid layer, given by $i_t$ in the figure, called the input gate, computes which values would be updated,followed by a tanh layer creating new candidate values, $\tilde{C_t}$, that can be added to the state $t$. Then $i_t$ and $\tilde{C_t}$ values are used to compute new cell state value $C_t$ as shown in the figure below. Finally, the Long short-term memory (LSTM) network computes output value $o_t$ for time step $t$ using $(h_{t-1},x_t)$. Then, the cell state is passed through tanh to transform the values in the range $[-1,1]$ and multiply it by $o_t$ to pass on as the history to the next time step $t$. This architecture makes LSTMs highly effective in natural language processing, machine translation, image processing, and reinforcement learning (\cite{LSTMreview}).
\end{doublespace}
\end{document}